\documentclass[runningheads]{llncs}
\usepackage[T1]{fontenc}
\usepackage{graphicx}
\usepackage{makecell}
\usepackage{multicol}
\usepackage{amsmath}
\usepackage{amsfonts}
\usepackage{amssymb}
\usepackage{algorithm}
\usepackage[noend]{algorithmic}
\usepackage{caption}
\usepackage{subcaption}
\DeclareMathAlphabet{\pazocal}{OMS}{zplm}{m}{n}
\newcommand{\stt}[1]{{\small\texttt{#1}}}

\begin{document}
\title{Online inductive learning from answer sets for efficient reinforcement learning exploration}
%
%
%
\author{Celeste Veronese\inst{1}\orcidID{0009-0007-7461-4039} \and
Daniele Meli\inst{1}\orcidID{0000-0002-3162-388X} \and
Alessandro Farinelli\inst{3}\orcidID{0000-0002-2592-5814}}
\authorrunning{C. Veronese et al.}
%
\institute{University of Verona, Department of Computer Science, Verona, Italy
\email{\{celeste.veronese,daniele.meli,alessandro.farinelli\}@univr.it}}
\maketitle              
\begin{abstract}
This paper presents a novel approach combining inductive logic programming with reinforcement learning to improve training performance and explainability. We exploit inductive learning of answer set programs from noisy examples to learn a set of logical rules representing an explainable approximation of the agent’s policy at each batch of experience. We then perform answer set reasoning on the learned rules to guide the exploration of the learning agent at the next batch, without requiring inefficient reward shaping and preserving optimality with soft bias. The entire procedure is conducted during the online execution of the reinforcement learning algorithm. We preliminarily validate the efficacy of our approach by integrating it into the Q-learning algorithm for the Pac-Man scenario in two maps of increasing complexity. Our methodology produces a significant boost in the discounted return achieved by the agent, even in the first batches of training. Moreover, inductive learning does not compromise the computational time required by Q-learning and learned rules quickly converge to an explanation of the agent’s policy.

\end{abstract}
\section{Introduction}
The intersection of symbolic, logic-based reasoning and Reinforcement Learning (RL) represents a trending topic in Artificial Intelligence (AI) research, aiming to harness the strengths of both domains and address their individual limitations \cite{Cheng2021HeuristicGuidedRL}. Current state-of-the-art approaches in model-free RL, in fact, predominantly rely on extensive data or pre-defined environmental models, facing significant challenges in terms of efficiency and scalability. Furthermore, the decision process underlying policy generation is not transparent to other agents and humans, hindering certifiability and trustworthiness.
On the other hand, symbolic AI works well in the small data regime but does not perform well on non-symbolic data and is not noise tolerant \cite{Vermeulen2023AnEO}.
Incorporating symbolic and logical formalisms into RL systems, as highlighted by \cite{Kambhampati2022Symbols}, can significantly boost their interpretability, thereby fostering wider acceptance and diffusion while also playing a crucial role in improving the policy and the training phase. However, symbolic learning and reasoning may significantly increase the computational burden of RL, thus limiting the effective real-time integration of neurosymbolic approaches \cite{FurelosBlanco2021InductionExploitation} or the scalability to non-trivial tasks, e.g., where multiple predicates and scopes for logical variables shall be defined \cite{marra2021neural}.

Inspired by results obtained by \cite{Mazzi2023LearningLogic,Meli2024LearningLogic} in model-based RL, we present a novel approach that exploits Inductive Logic Programming (ILP, \cite{Muggleton1991InductiveLogicProgramming}) to iteratively learn human-interpretable policy heuristics \emph{online}, during the training of RL agents, and directly use them to bias the agent in its exploration process.
Specifically, for each batch of experiences (state-action sequences) collected by the RL agent during training, we rank them by the cumulative return and convert them to a logical formalism to build examples for ILP. This representation maps states and actions to basic human-level concepts about the task, thus enhancing the interpretability of the trained policy. 
ILP then learns a logical approximation of the policy, which is then used in the following batch to bias the agent's exploration to improve the performance by avoiding wrong or dangerous actions.
We adopt the logical formalism of Answer Set Programming (ASP) \cite{Lifschitz1999AnswerSetPlanning} to represent logical specifications and perform online heuristic reasoning. ASP is chosen as it can be considered the state-of-the-art in planning domain representation for autonomous agents \cite{meli2023logic}.
In this way, we can rely on the latest developments in scalable and efficient ILP, even in the presence of many examples \cite{Law2020FastLAS}. 
We preliminarily apply our methodology on the simple approximate Q-learning algorithm \cite{SuttonBarto2018}, to highlight better the variations introduced by our approach. However, any RL algorithm that includes a random exploration phase could be used (e.g. Double Deep Q-Network \cite{Hasselt2015DeepRL}).
In summary, this paper presents the following contributions to the state-of-the-art:
\begin{itemize}
\item we introduce an innovative approach to online neurosymbolic learning and reasoning. Our methodology does not simply shape the reward of the agent according to the learned logical heuristics, but directly guides RL exploration, resulting in a tighter and more efficient neurosymbolic interconnection \cite{Cheng2021HeuristicGuidedRL}. Moreover, we preserve RL optimality guarantees adopting a probabilistic soft bias approach; 
\item differently from \cite{Meli2024LearningLogic}, we perform the learning of informative logical heuristics \emph{online} during RL training while maintaining the expressiveness of ASP. To this aim, we exploit fast scalable ILP as in \cite{Law2020FastLAS};
\item we empirically evaluate the effectiveness of our methodology in the benchmark Pac-Man scenario on maps of increasing complexity with contrastive agents. We obtain a significant boost in the discounted return achieved by the agent. Furthermore, while the inductive learning component minimally affects computational time, it facilitates a swift convergence of learned rules, leading to a coherent and comprehensive explanation of the underlying black-box policy.
\end{itemize}

\section{Related Work}
Recent works demonstrate that incorporating existing knowledge into RL and Markov Decision Processes (MDPs) can greatly enhance the development of effective policies \cite{Cheng2021HeuristicGuidedRL}.
The integration of such knowledge through logical formalisms has significantly improved policy computation. For instance, the REBA framework by \cite{Sridharan2019REBA} employs ASP to define spatial relations in a domestic setting, guiding a robotic agent to select specific rooms for inspection while addressing more straightforward MDP tasks locally. Similarly, DARLING \cite{Leonetti2016Synthesis} uses ASP to restrict MDP exploration in simulated grid environments and real-world robotics applications. Furthermore, \cite{DeGiacomo2019RestrainingBolts} and \cite{Leonetti2012AutomaticGeneration} utilize linear temporal logic to direct exploration in MDPs. 
Logical constraints also play a crucial role in avoiding undesirable behaviours, particularly in safety-sensitive scenarios \cite{Mazzi2023RiskAwareShielding,Marzari2023OnlineSafety}. However, since logical heuristics usually express policy-related knowledge (unknown to the agent), they need to be defined by expert users. In the alternative, a huge amount of high-quality training examples are needed to learn them, such as in \cite{DeGiacomo2020ImitationLearning}, in which temporal logic specifications are learnt as finite automata from good example traces and used to shape the reward signal. 
For this reason, attempts have been made to combine neural and symbolic learning.
For instance, in \cite{marra2021neural} Neural Markov Logic Networks are used to learn relational representations from structured examples. However, the approach is computationally inefficient out of very simple domains, involving few variables and predicates. 
In \cite{hazra2023deep} a deep relational reinforcement learning approach to learn generalizable policies is proposed. Nonetheless, the agent only adopts the logical policy at training convergence, thus not fully exploiting the synergy with neural methods, which are inherently more noise-robust and can help refine the agent's performance. Furthermore, even the solution presented in \cite{hazra2023deep} suffers from poor scalability, requiring task-specific constraints on the available search space for logical policies.

In this paper, we combine deep RL with an ILP framework under the ASP semantics, which offers great expressiveness for structured task representation and reasoning in a fragment of first-order logic \cite{meli2023logic}.
ILP under the ASP semantics has been proven successful in numerous scenarios, e.g., in enhancing the explainability of black-box models \cite{Veronese2023Inductive,Rabold2018ExplainingBlackBox} and gaining task knowledge \cite{Rodriguez2021LearningFirstOrder,Meli2021InductiveLearning}.
Among different available implementations \cite{Hocquette2021Complete,Cropper2021Learning}, we adopt the popular FastLAS system \cite{Law2020FastLAS}, which can scale to very large search spaces.
In this way, we can efficiently learn ASP policy approximations online while training the RL agent, solving the computational limitations posed by \cite{marra2021neural,hazra2023deep}.

Our work is close to the one in \cite{FurelosBlanco2021InductionExploitation}, where ASP rules are learned to define reward machines in RL settings. However, our solution is more scalable thanks to the FastLAS approach. Furthermore, we do not use rules to shape the reward signal of the RL agent, which may be inefficient in effectively boosting the performance of RL training since still many interactions with the environment may be required to learn the value of action \cite{Cheng2021HeuristicGuidedRL}. Instead, we take inspiration from \cite{Meli2024LearningLogic} and perform ASP reasoning over the learned policy heuristics to bias the exploration of an RL agent with a higher probability towards heuristic-suggested actions.

\section{Background}
We now introduce some basic notions about the approximate Q-Learning algorithm, ASP and ILP, and describe the benchmark domain we used to test our methodology. 

\subsection{Approximate Q-Learning}
Q-learning is a model-free reinforcement learning algorithm used to find an optimal action-selection policy \cite{SuttonBarto2018}. The goal of the agent is to learn the optimal policy that maximizes the expected cumulative reward over time. The core concept in Q-learning is the action-value function, \( Q(s, a) \), which estimates the expected future rewards for taking action \(a\) in state \(s\).
The Q-learning update rule is given by:
\[
Q(s_t, a_t) \leftarrow Q(s_t, a_t) + \alpha \left( r_{t+1} + \gamma \max_{a'} Q(s_{t+1}, a') - Q(s_t, a_t) \right)
\]
where \(s_t\) is the current state, \(a_t\) is the action taken in that state, \(r_{t+1}\) is the reward received after taking action \(a_t\), and \(s_{t+1}\) is the next state. The parameter \(\alpha \in [0,1]\) is the learning rate, which controls how much new information overrides old information, and \(\gamma \in [0,1]\) is the discount factor, which determines the importance of future rewards.
The algorithm converges to the optimal action-value function \(Q^*(s, a)\) under certain conditions, such as a decaying learning rate and sufficient exploration of all state-action pairs. The optimal policy \(\pi^*(s)\) is then derived by choosing the action that maximizes the Q-value in each state:
\[
\pi^*(s) = \arg \max_{a} Q^*(s, a).
\] 
Approximate Q-learning addresses scalability issues inherent in traditional Q-learning, particularly in environments with large or continuous state spaces \cite{SuttonBarto2018}. The key idea is to approximate the Q-function using a combination of features instead of the full state. That is, instead of recording everything in detail, we think about what is most important to know, and model that. 

\subsection{Answer Set Programming}
Answer Set Programming (ASP) defines a domain as a set of logical statements (axioms) articulating the logical relationships between entities represented as predicates and variables (atoms) \cite{Gelfond1988TheSM}. 
Axioms considered in this work are \emph{normal rules} $\mathtt{h :\!\text{- } b_1,\ldots,b_n}$, which define the body of the rule (i.e. the logical conjunction of literals $\mathtt{\bigwedge^n_{i=1}b_i}$) as a precondition for the head $\mathtt{h}$. 
We say that a variable is grounded when assigned a particular value. Consequently, an atom is grounded when all its variables are grounded. Given an ASP program $P$, its Herbrand base $\pazocal{H}(P)$ defines the set of ground atoms which can be generated from it. From an ASP domain definition, an ASP solver computes the \emph{answer sets}, i.e., the minimal set of literals that satisfies the given logic program according to the stable model semantics. This work assumes that body atoms represent environmental features describing $S$ in the MDP, while head atoms are actions. 
Answer sets will then contain feasible actions for the RL agent.

\subsection{Inductive Logic Programming}
A generic ILP task under a logical formalism $F$ \cite{Muggleton1991InductiveLogicProgramming} is defined as a tuple $\mathcal{T}= \langle B, S_M, E \rangle$, consisting of background knowledge $B$ expressed in a logic formalism $F$, a search space $S_M$ containing the set of possible axioms to be learned (defined, e.g., via a mode declaration $M$ \cite{muggleton1995inverse}), and a set of examples $E$, all expressed in the syntax of $F$. The goal is to find a hypothesis (i.e. an axiom) $H \subseteq S_M$ covering $E$. 
Under the ASP semantics, we consider the generic case where examples are \emph{weighted context dependent
partial interpretations} (WCDPI’s) \cite{Law2018InductiveLearningASP}. A partial interpretation $e^{pi}$ is a pair of sets of ground atoms $\langle e^{inc}, e^{exc} \rangle$. 
An interpretation (i.e., a set of ground atoms) $I$ extends $e$ iff $e^{inc} \subseteq I$ and $e^{exc} \cap I = \emptyset$. A WCDPI is then a tuple $e = \langle e^{id}, e^{pen}, e^{pi}, e^{ctx} \rangle$, where $e^{id}$ is an identifier for $e$, $e^{pen}$ is either a positive integer or $\infty$, called a penalty, $e^{pi}$ is a partial interpretation, and $e^{ctx}$ is an ASP program called the context. A WCDPI $e$ is accepted by a program $P$ iff there is an answer set of $P \cup e^{ctx}$ that extends $e^{pi}$.
Following the definition by \cite{Law2018InductiveLearningASP}, the goal of ILP is then to find a hypothesis $H \subseteq S_M$ with minimal length (i.e., number of atoms) and $\sum_{e}e^{pen}$, for all examples $e$ not accepted by $H\cup B$.
Since we are interested in discovering normal rules matching actions to environmental features, $e^{inc}, e^{exc}$ represent executed and not executed actions, respectively, while $e^{ctx}$ is the set of ground environmental features.
We employ FastLAS learner by \cite{Law2020FastLAS} for fast computation from acquired examples gathered from RL batches of experience.

\subsection{The Pac-Man domain}
\begin{figure}
\centering
\includegraphics[width=0.3\linewidth]{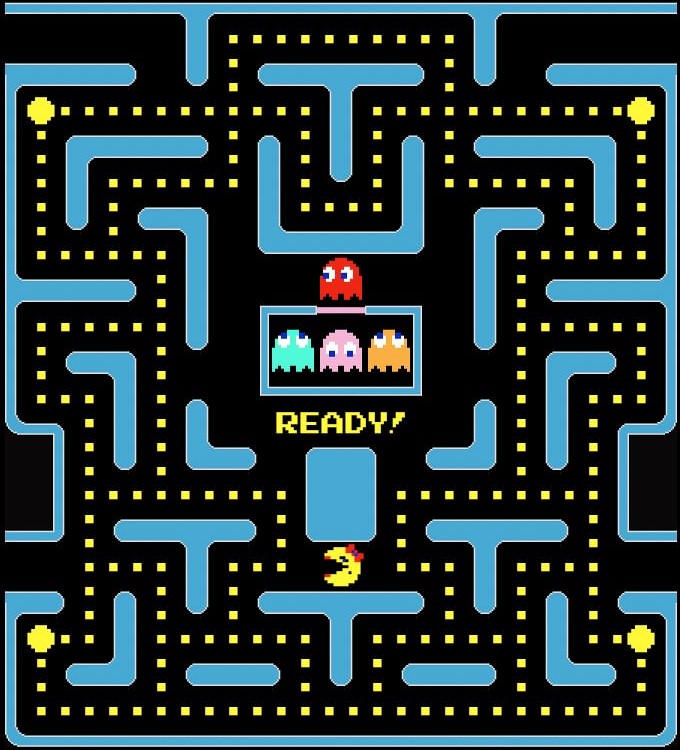}
\caption{Example scenario for the Pac-Man domain, $G=P=4$, $25\times26$ grid.}\label{fig:pacman}
\end{figure}

In the Pac-Man domain, an agent (Pac-Man) needs to navigate a maze-like environment to collect food pellets (each one gives a +10 reward) while avoiding enemies (ghosts). The $G$ ghosts present in the environment generally move randomly but may start chasing Pac-Man (with probability $p_g$) if they are close (we conducted our experiments with $p_g=0.8$). Given the $N\times M$ grid, the agent can move in the four cardinal directions or stay still (hence $|A|=5$), receiving a time penalty of -1 for each time step spent in the maze. The episode ends with a +500 reward if Pac-Man eats all food pellets or a -500 penalty if one of the ghosts chases Pac-Man. The environment also contains $P$ power capsules (big yellow dots in Figure \ref{fig:pacman}) that Pac-Man can eat to gain the ability to scare and eat ghosts. Catching a scared ghost results in a +200 reward. The state $S$ contains the position of each wall, food pellet, power pill, ghost in the map, and the Pac-Man position; each is expressed as $(x, y)$ coordinates. The environment is fully deterministic.
The challenge Pac-Man presents derives from the vast dimensions of the state space and the extended planning horizon: to complete the level, the agent may have to explore the entire environment to find all the pellets, and non-trivial movements are often required to escape ghosts.

\section{Methodology}
This section describes our methodology for integrating approximate Q-learning, symbolic learning, and reasoning in ASP. We exemplify it in the context of the standard RL Pac-Man domain used for empirical evaluation. 

\subsection{ASP representation of the domain}
We start from the representation of the domain in ASP syntax. This requires defining environmental features $\pazocal{F}$ and actions $\pazocal{A}$.
For the Pac-Man domain, $\pazocal{F}$ contains: \stt{wall(Dir)}, which denotes the presence of a wall in front of the agent (i.e., one cell away) in that direction, \stt{food(Dir, Dist)}, \stt{ghost(Dir, Dist)}, and \stt{capsule(Dir, Dist)}, representing Manhattan distance \stt{Dist} $\in [0, 10]$ between PacMan and a cell containing food (or ghost, or capsule) in direction \stt{Dir} $\in$ \stt{\{north, south, east, west\}}. Finally, we need to introduce upper and lower bounds on \stt{Dist} to obtain more informative policy heuristics. 
To this aim, we define atoms in the form \stt{X\_dist\_Y(Dir,Dist,D)}, where \stt{X} is either \stt{ghost, food} or \stt{caps}, and \stt{Y} is either \stt{geq} or \stt{leq}, defined as follows:
\begin{align*}
    &\stt{X\_dist\_geq(Dir, Dist, D) :- X(Dir, Dist), Dist >= D, d\_const(D).} \\  
    &\stt{X\_dist\_leq(Dir, Dist, D) :- X(Dir, Dist), Dist <= D, d\_const(D).}
\end{align*} 
where \stt{d\_const(0..4)} limits the set of possible values that \stt{D} can take in the rule.
Action atoms are constructed from $A$ as \stt{move(Dir)}, denoting the movement of the agent in direction \stt{Dir}. The agent also has the option to perform a 'stop' action. However, we chose not to include it in the learning phase as it was rarely performed in the examples collected during our tests.
Once $\pazocal{F}$ and $\pazocal{A}$ are defined, we need a \emph{feature map} $F_\pazocal{F} : S \rightarrow \pazocal{H(F)}$ and an \emph{action map} $F_\pazocal{A} : A \rightarrow \pazocal{H(A)}$ to ground atoms from collected batches of RL. 
The only information needed to build $\pazocal{F}$ is the positions of the agent, food, ghosts and capsules in the environment, all of which are available in $S$. 

\subsection{Definition of the learning task}
\label{subsec:ilp_met}
Given the ASP formalization of the task, we need to generate the ILP task $T = \langle B, S_M, E \rangle$ for FastLAS, starting from RL episodes.
Given an episode consisting of a sequence of $N$ state-action pairs $\langle \Bar{a}_i, \Bar{s}_i \rangle, \ i = 1, \ldots N$, we can build a WCDPI of the following form:
\[
e_i = \langle id_i, w_i, \langle \{\Bar{a}_i\}, a\in F_{\pazocal{A}} \setminus F_{\pazocal{A}}(\Bar{a}_i) \rangle, F_{\pazocal{F}}(\Bar{s}_i) \rangle,
\]
where $id_i$ is a unique identifier and $w_i$ represents the penalty of the WCDPI, set as the reward obtained by the agent in that episode and assigning $w_i = 0$ to those episodes that obtained a negative reward.\footnote{This is necessary because FastLAS does not handle negative weights. However, as explained in the following section, we only consider the best episodes of each batch; thus, the presence of rewards less than zero is highly improbable.} This gives more relevance to the agent's behaviour, resulting in a higher reward for generating rules that can lead to at least the same performance. 
We populate $e^{inc}$ with the agent's chosen action and $e^{exc}$ with the grounding for all unobserved actions. 
For example, considering the Pac-Man scenario depicted in Figure \ref{fig:pacman}, if at time $t$ the agent moves left within an episode with return $50$, we generate the following WCDPI:
\begin{align}
e_t = &\langle id_t, 50, \langle\stt{move(east)}, \{ \stt{move(west)}, \stt{move(north)}, \stt{move(south)} \} \rangle, \nonumber\\ &\{ \stt{wall(north)}, \stt{wall(south)}, \stt{food(east, 1)}, \stt{food(west, 1)},\ldots\} \rangle,
\end{align}
We have omitted ground context atoms with a distance higher than 1 for simplicity.
The background knowledge of the task only contains the definition of the ASP variables and ranges, and the mode declaration is defined in order only to have rule's heads in the form \stt{move(Dir)}, and bodies containing atoms from $\pazocal{F}$. 

\subsection{Neurosymbolic Q-learning}
Algorithm \ref{alg:approx_q_learning} shows our Neurosymbolic Q-Learning methodology.
At the end of each batch of experience acquired by Approximate Q-Learning, we store at most $\sigma$
highest-return episodes (Line 22). We then follow the procedure described in Section \ref{subsec:ilp_met} to generate WCDPIs and learn policy heuristics $H$ (Line 23). These heuristics, together with the background knowledge $B$ defined in Section \ref{subsec:ilp_met} and the grounding of state variables $F_{\pazocal{F}}(s)$, are used to perform ASP reasoning and compute a set of \emph{suggested actions} $A_h$ at each step of the Q-learner (Line 7). Specifically, in the exploration phase of the agent (with probability $\epsilon$), it may choose either an action from $A_h$ or $A \setminus A_h$ (i.e., actions which $ H$ does not suggest), with probability $\rho$ and $1-\rho$ respectively (Line 10). In this way, we preserve the asymptotic optimality of the Q-learning algorithm \cite{SuttonBarto2018}, following a soft bias approach as in existing literature \cite{Meli2024LearningLogic}. We empirically choose $\rho$ as the average discounted return over the best-saved episodes $E_{\sigma}$, normalized by the average return of the whole training. 
In order not to start with $A_h = \emptyset$ until the first batch of episodes is gathered, we use the very first episode of training to generate preliminary policy heuristics (Line 21), which provides a greedy yet useful hint to the agent, as explained in Section \ref{sec:exp}.
We remark that the described methodology can be equivalently applied to any RL algorithm, where random actions are taken in the exploration phase \cite{SuttonBarto2018}.

\begin{algorithm}
\caption{Neurosymbolic Q-Learning}
\label{alg:approx_q_learning}
\textbf{Require}: MDP = $\langle S, A, R, T, \gamma \rangle$, background knowledge $B$, search space $S_M$\\
\textbf{Parameters}: Q-learning parameters \cite{SuttonBarto2018} (including $\epsilon \in [0, 1]$), max best episodes stored $\sigma$, max episodes $E$, batch size $S_b$\\
\begin{algorithmic}[1] 
\STATE Initialize weight vector $\mathbf{w}$, batch $b = \emptyset$, episode $e = \emptyset$, episode count $N_e=0$, policy heuristics $H = \emptyset$, set of best episodes $E_{\sigma} = \emptyset$
\WHILE{$N_e < E$}
    \STATE Observe initial state $s$
    \WHILE{$s$ is not terminal}
        \STATE $x \sim [0,1]$
        \IF{$x < \epsilon$}
            \STATE $A_h$ $\gets$ \textbf{ASP($B, H, F_\pazocal{F}(s)$)} 
            \IF{$A_h \neq \emptyset$}
                \STATE $\rho \gets$ \textbf{AvgReturn}($E_{\sigma}$)
                \STATE $a \sim$ \textbf{WeightedProb($A, F^{-1}_{\pazocal{A}}(A_h), \rho$)} 
            \ELSE
                \STATE $a \sim A$
            \ENDIF
        \ELSE
            \STATE $a \gets \arg\max_{a'} Q(\mathbf{w}, s, a')$
        \ENDIF
        \STATE $e$.Append($\langle s, a \rangle$)
        \STATE $s' \gets T(s,a), \ r \gets R(s,a,s')$
        \STATE $\mathbf{w} \gets$ \textbf{QUpdate($s,a,r,s', \gamma$)}
        \STATE $s \gets s'$
    \ENDWHILE
    \STATE $N_e \gets N_e+1$
    \STATE $b.$\textbf{Append($e$)}
    \IF{$|b| = S_b \lor N_e = 1$}
        \STATE $E_{\sigma} \gets$ \textbf{Update($b, \sigma$)}
        \STATE $H \gets$ \textbf{FastLAS($B, S_M, E_\sigma$)} \hfill\COMMENT{$E_\sigma$ converted to ASP syntax via $F_{\pazocal{F}}, F_{\pazocal{A}}$}
    \ENDIF
\ENDWHILE
\end{algorithmic}
\end{algorithm}

\section{Empirical Evaluation}
\label{sec:exp}
Experimental results on the methodology outlined in the previous section are now presented. We tested our approach on two different scenarios of the Pac-Man domain: a smaller map, with $18\times9$ grid dimensions, $P=2$ power capsules and $G=2$ ghosts, and a way more challenging map, with $25\times26$ grid dimensions, $G=4$ ghosts and $P=4$ power capsules.
All experiments were performed on a computer equipped with 5.1GHz, a 13th Gen Intel i5 processor and 64GB RAM. 
We evaluate 2 different performance measures:
\begin{itemize}
    \item \emph{training performance}: it measures the RL discounted return and computational efficiency;
    \item \emph{policy heuristic convergence}: it measures the explainability and robustness of the symbolic component of our methodology, evaluating how the set of policy heuristics reaches a fixed point as RL training progresses.
\end{itemize}
For all metrics, we compare our methodology (\textbf{NeuroQ}) against classical Approximate Q-Learning (\textbf{ApproxQ}).
We report results as mean and standard deviation over a set of 5 random seeds for statistical relevance.
For each seed, we choose a maximum number of episodes $E=20000$ and a batch size $S_b = 100$ (resulting in 200 training batches).
Q-Learning parameters for both evaluated algorithms are set as follows: learning rate $\alpha = 0.2, \gamma = 0.8, \epsilon = 0.05$. We empirically choose to keep track of $\sigma = 5$ best episodes on the smaller map, while on the bigger map (in which longer episodes are generated), we set $\sigma = 3$. In this way, we balance RL performance (discounted return) and the computational cost of symbolic learning.

\subsection{Training performance}
\begin{figure}
\begin{subfigure}[b]{0.49\linewidth}
\includegraphics[width=\linewidth]{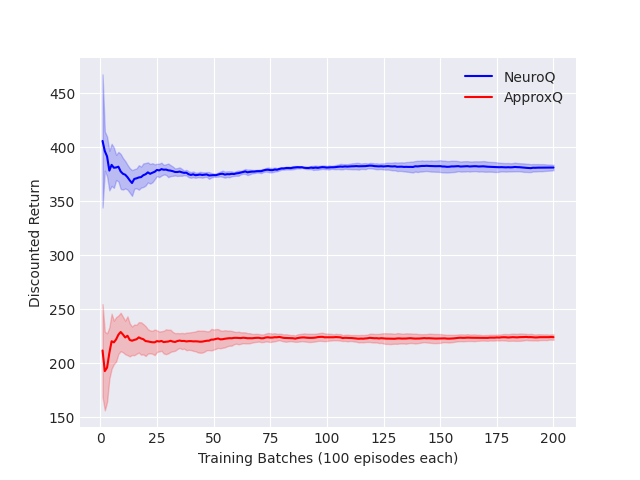}
\caption{$18\times9$ map, $G=P=2$.}\label{fig:reward_medium}
\end{subfigure}
\hfill
\begin{subfigure}[b]{0.49\linewidth}
\includegraphics[width=\linewidth]{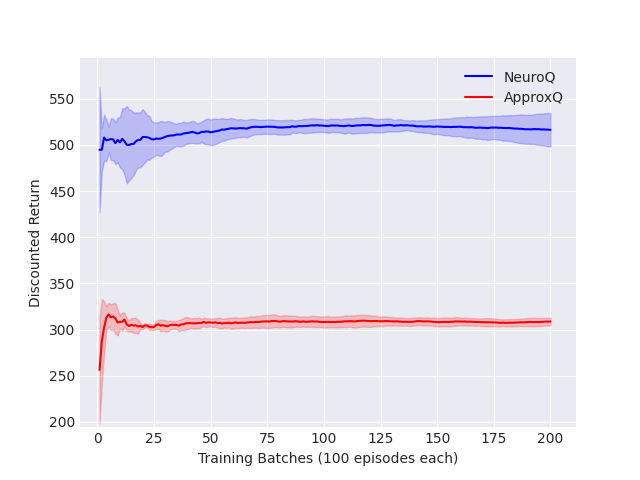}
\caption{$25\times26$ map, $G=P=4$.}\label{fig:reward_original}
\end{subfigure}
\caption{Average discounted return in the Pac-Man domain. 
}\label{fig:reward}
\end{figure}

Figure \ref{fig:reward} shows the return obtained by the execution of ApproxQ and NeuroQ in the small map (\ref{fig:reward_medium}) and in the more challenging map (\ref{fig:reward_original}). We note that, in both scenarios, the introduction of policy heuristics significantly increases the return obtained by the agent, leading to a more efficient training process.
In addition to assessing the trade-off between performance and computation, we investigated the computational impact of generating policy heuristics and calculating the suggested actions. 
The execution times, shown in Tables \ref{tab:training_times_medium}-\ref{tab:training_times_original} for the small and large maps, respectively, demonstrate that the additional time required by FastLAS learning and ASP reasoning (Lines 7 and 23 in Algorithm \ref{alg:approx_q_learning}) is acceptable, given the substantial increase of performance. In particular, in the large map, NeuroQ requires $\approx 25\%$ more time per batch, but the average discounted return is double (Figure \ref{fig:reward_original}).

\setlength{\tabcolsep}{5pt}
\begin{table}[ht]
\centering
\caption{Execution times (in seconds) in total and per batch, on the $18\times9$ map with $G=P=2$. Mean and standard deviation (in brackets) are reported where appropriate.}
\label{tab:training_times_medium}
\vspace{3mm}
\resizebox{0.9\linewidth}{!}{
\begin{tabular}{ccccc}
\hline
Seed & \multicolumn{2}{c}{ApproxQ} & \multicolumn{2}{c}{NeuroQ} \\
& Total & Per batch & Total & Per batch\\ 
\hline 
0 & 949.39 & 4.74 ($\pm 0.67$) & 1505.91 & 7.53 ($\pm 0.57$) \\
1 & 955.86 & 4.78 ($\pm 0.65$) & 1497.01 & 7.48 ($\pm 0.60$) \\
2 & 953.72 & 4.77 ($\pm 0.67$) & 1508.62 & 7.54 ($\pm 0.56$) \\
3 & 952.75 & 4.76 ($\pm 0.68$) & 1496.73 & 7.48 ($\pm 0.57$) \\ 
4 & 953.46 & 4.77 ($\pm 0.67$) & 1503.07 & 7.52 ($\pm 0.60$) \\ 
Average & 953.04 ($\pm 2.34$) & 4.77 ($\pm 0.67$) & 1502.27 ($\pm 4.92$) & 7.51 ($\pm 0.58$)\\
\hline
\end{tabular}
}
\end{table}

\begin{table}[ht]
\centering
\caption{Execution times (in seconds) in total and per batch, on the $25\times26$ map with $G=P=4$. Mean and standard deviation (in brackets) are reported where appropriate.}
\label{tab:training_times_original}
\vspace{3mm}
\resizebox{0.9\linewidth}{!}{
\begin{tabular}{ccccc}
\hline
Seed & \multicolumn{2}{c}{ApproxQ} & \multicolumn{2}{c}{NeuroQ} \\
& Total & Per batch & Total & Per batch\\ 
\hline 
0 & 3980.74 & 19.9 ($\pm 3.26$) & 4974.46 & 24.8 ($\pm 3.44$) \\
1 & 4005.92 & 20.0 ($\pm 3.46$) & 5233.66 & 26.1 ($\pm 3.20$) \\
2 & 4054.08 & 20.2 ($\pm 3.37$) & 5197.47 & 26.0 ($\pm 3.16$) \\
3 & 4003.02 & 20.0 ($\pm 3.45$) & 5196.34 & 26.0 ($\pm 3.26$) \\ 
4 & 4025.39 & 20.1 ($\pm 3.13$) & 5174.59 & 25.9 ($\pm 3.35$) \\ 
Average & 4013.83 ($\pm 27.51$) & 20.07 ($\pm 3.33$) & 5155.30 ($\pm 105.7$) & 25.78 ($\pm 3.28$)\\
\hline
\end{tabular}
}
\end{table}

\subsection{Policy heuristics convergence}

At each batch iteration of Algorithm \ref{alg:approx_q_learning}, the ILP task solved by FastLAS consists of $\approx 650$ WCDPIs (on average) for both the small and large maps. The search space $S_M$ remains constant during the procedure and consists of 226 unique rules for both environments.
As the first training episode is acquired (Line 21 of Algorithm \ref{alg:approx_q_learning}), we learn a first set of policy heuristics to bias the agent's exploration towards more convenient actions immediately. On the $18\times9$ map, for example, the following heuristic is generated: 
\[
\stt{move(Dir) :- food\_dist\_leq(Dir,Dist,1).}
\]
In other words, the agent immediately learns to move in the direction where the food is close. 
In the $25\times 26$ map, FastLAS learns a similar heuristic at the first iteration, with slight adjustments depending on the specific seed (e.g., \stt{not wall(Dir)} is included in the body, too).

As the training progresses, learned heuristics finally converge to the following in the smaller environment: 
\begin{subequations}
\begin{align}
    \stt{move(Dir) :- } &\stt{food\_dist\_leq(Dir,Dist,1).} \\
    \stt{move(Dir) :- } &\stt{caps\_dist\_leq(Dir,Dist,1).}
\end{align}
\end{subequations}
The first greedy heuristic learned in the first episode is confirmed, while the second rule pushes the agent towards close power capsules.
On the $25\times26$ map, the final set of learned rules is the following:
\begin{subequations}
\begin{align}
    \stt{move(Dir) :- } &\stt{caps\_dist\_leq(Dir,Dist,1).} \\
    \stt{move(Dir) :- } &\stt{food\_dist\_leq(Dir,Dist,1).} \\
    \stt{move(Dir) :- } &\stt{not wall(Dir), } \stt{food\_dist\_leq(Dir,Dist,2).} \\
    \stt{move(Dir) :- } &\stt{ghost\_dist\_geq(Dir,Dist,4).}
\end{align}
\end{subequations}
In this case, the agent interacts with a more complex map for a longer time. As a consequence, the policy heuristics are more informative and also capture the necessity of avoiding ghosts and walls, stating that the agent should move in a direction only if no walls are present (\stt{not wall(Dir)}) or if ghosts are at a sufficiently high distance (\stt{ghost\_dist\_geq(Dir,Dist,4)}) in that direction. 

\begin{figure}
\begin{subfigure}[b]{0.49\linewidth}
\includegraphics[width=\linewidth]{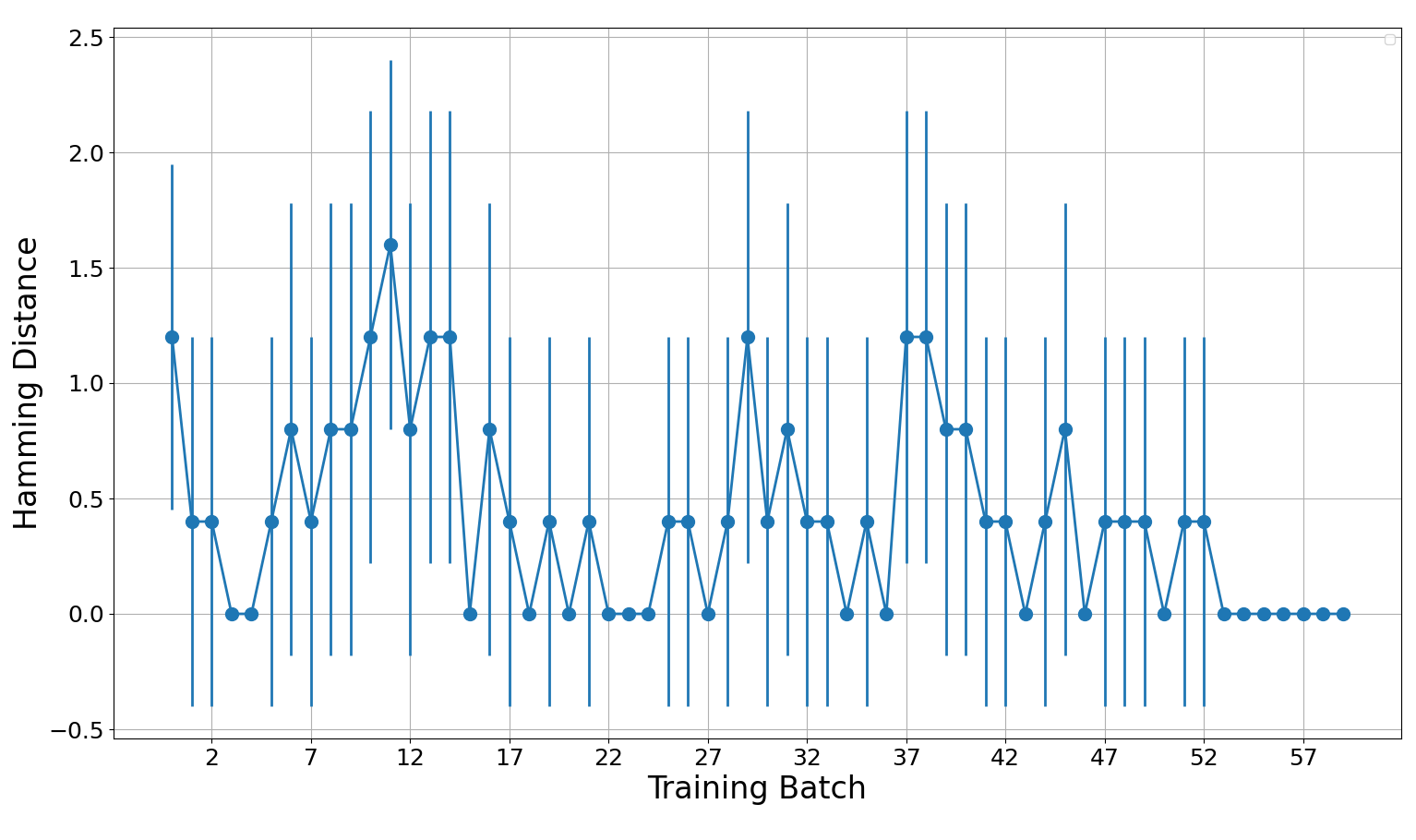}
\caption{$18\times9$ map, $G=P=2$.}\label{fig:hamming_medium}
\end{subfigure}
\hfill
\begin{subfigure}[b]{0.49\linewidth}
\includegraphics[width=\linewidth]{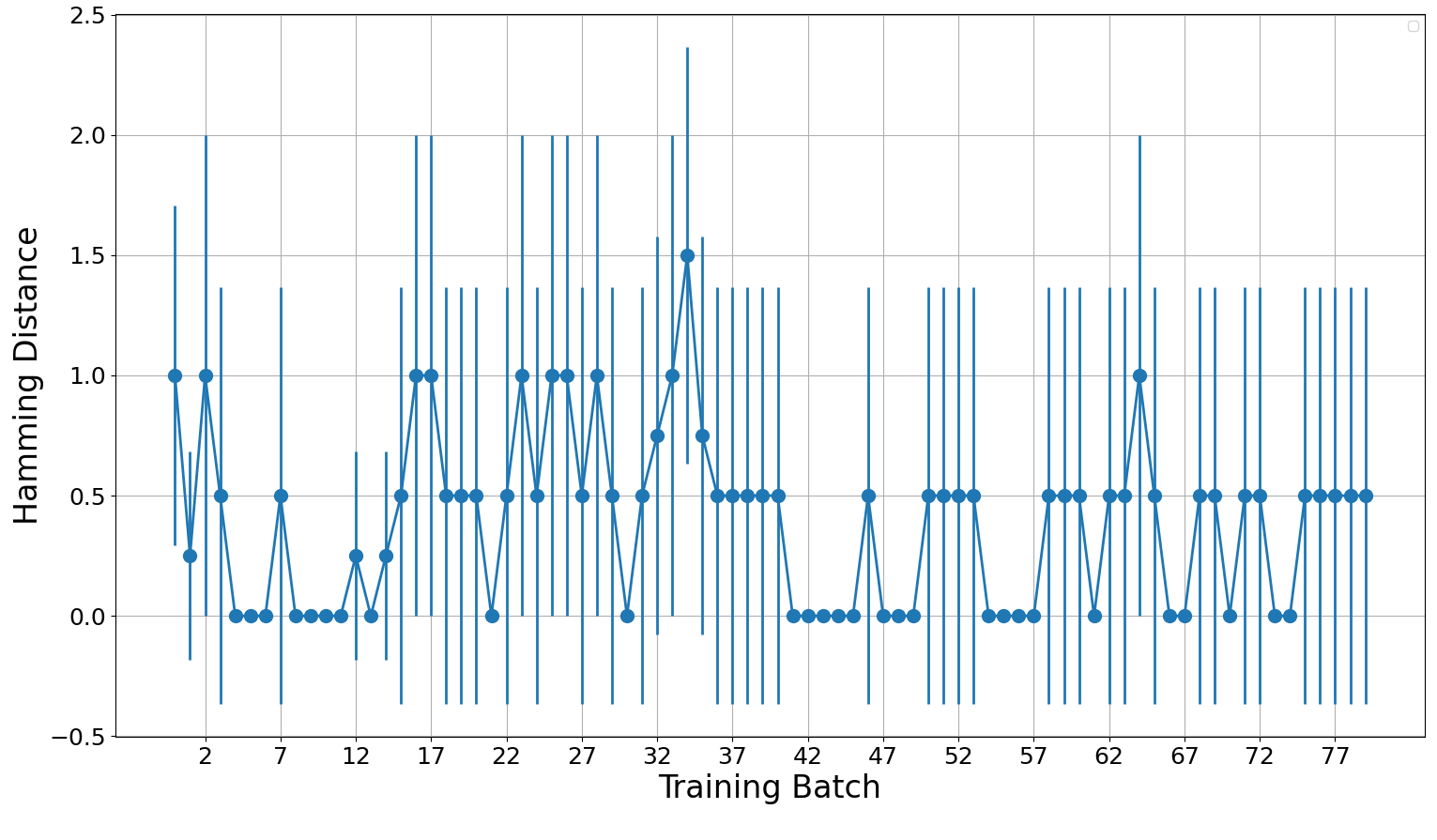}
\caption{$25\times26$ map, $G=P=4$.}\label{fig:hamming_original}
\end{subfigure}
\caption{Convergence of learned policy heuristics (mean $\pm$ standard deviation over seeds). To facilitate visualization, batches beyond convergence are omitted.}\label{fig:hamming}
\end{figure}

Figure \ref{fig:hamming} shows the convergence of the hypothesis for the small (\ref{fig:hamming_medium}) and large (\ref{fig:hamming_original}) maps, respectively. Convergence per batch is measured as the Hamming distance between the literals composing the body of rules extracted at a given batch, normalized by the number of literals composing the final sets of rules.
Ideally, we aim for zeroing Hamming distance when convergence is achieved.
The charts show that the symbolic learner converges within $< 70$ batches (over 200) in both scenarios, reflecting the convergence of the overall RL policy.
In particular, on the large map
(Figure \ref{fig:hamming_original}) rules don't converge to a zeroed Hamming distance (averaging over 5 seeds). Indeed, with one seed, FastLAS alternatively learns two sets of rules, one in which rule 3b is present and one in which it is not. However, rule 3c is always present and is more informative than 3b. Hence, this doesn't represent a significant change in the semantics of the policy.

\section{Conclusion and Future Work}
The research presented in this paper demonstrates how to effectively integrate neurosymbolic learning and reasoning to improve the efficiency and explainability of RL agents. 
We leverage the expressive ASP semantics to represent structured task knowledge in a fragment of first-order logic and reason over the most convenient actions in the RL exploration phase. ASP policy heuristics are learned and refined online via a scalable ILP algorithm, gathering examples at each batch of RL training.
Moreover, a probabilistic soft bias approach in ASP guidance preserves the convergence guarantees of RL.
We validated our algorithm in the Pac-Man benchmark RL scenario in two maps of increasing complexity involving long planning horizons, large state space and contrastive agents (ghosts). Our neurosymbolic RL algorithm extends Q-Learning, but it could be applied to any RL algorithm involving a random exploration phase.
Our methodology achieves a significantly higher discounted return than traditional RL (almost double in the largest map), at the cost of $\approx 25\%$ increase in the computational cost. 
In addition, learned ASP rules converge within $\approx <70/200$ RL batches, providing an interpretable logical explanation of the black-box RL policy and training process.
As a future work, we plan to test the scalability of these techniques to even more complex and varied environments, also investigating the integration with other RL algorithms and more complex forms of neurosymbolic integration \cite{Cheng2021HeuristicGuidedRL}.

\bibliographystyle{splncs04}
\bibliography{bibliography}

\end{document}